\documentclass[11pt]{article}

\usepackage[final]{style/acl}

\usepackage{times}
\usepackage{booktabs,tabularx,array}
\usepackage{latexsym}
\usepackage{enumitem}
\newcommand{\methodname}{\textit{HyperTrace }}
\newcommand{\method}{\textit{HyperTrace}}

\usepackage[T1]{fontenc}

\usepackage[utf8]{inputenc}

\usepackage{microtype}

\usepackage{inconsolata}

\usepackage{graphicx}
\usepackage{subcaption}
\usepackage{booktabs}

\title{HyperTrace: Hypothesis-Based Preference Tracing for Online \\ LLM Personalization}

\author{
  \textbf{Jianzhi Shen\textsuperscript{1,*}},
  \textbf{Keyu Mao\textsuperscript{2,*}},
  \textbf{Minghao Shao\textsuperscript{3,4}},
  \textbf{Chuanyang Jin\textsuperscript{1}},
\\
  \textbf{Yusong Wang\textsuperscript{2}},
  \textbf{Ailiang Lin\textsuperscript{2}},
  \textbf{Kotaro Funakoshi\textsuperscript{2}},
  \textbf{Manabu Okumura\textsuperscript{2}},
\\
\textbf{Tianmin Shu\textsuperscript{1}},
  \textbf{Muhammad Shafique\textsuperscript{4,$\dagger$}}
\\
\\
  \textsuperscript{1}Johns Hopkins University,
  \textsuperscript{2}Institute of Science Tokyo,
  \textsuperscript{3}NYU Tandon,
  \textsuperscript{4}NYU Abu Dhabi
\\
  \small{
    \textsuperscript{*}Equal contribution.
    \textsuperscript{$\dagger$}Corresponding Author.
    \textbf{Correspondence:}
    \href{mailto:muhammad.shafique@nyu.edu}{muhammad.shafique@nyu.edu}
  }
}

\begin{document}

\maketitle

\begin{abstract}
Personalized language models aim to adapt responses to individual users, whose preferences are often latent and revealed gradually through interaction.
Existing training-free methods rely on stored histories or retrieved memories, but they often struggle to reconcile long-term preferences with short-term topic-specific needs.
To address this issue, we propose \method, a training-free framework that formulates online personalization as latent preference tracing.
\methodname maintains interpretable natural-language hypotheses over short-term intent and long-term preferences, and updates them through an SMC-style reweight process using an LLM-based surrogate choice model.
By updating these hypotheses across turns and sessions, \methodname enables personalization without parameter updates. Experiments on PRISM and PersonaMem-v2 show that \methodname improves response alignment, preference prediction, and profile consistency over strong online baselines, demonstrating the effectiveness of tracing latent user preferences for robust personalization. Code and scripts are available in the repository:
\url{https://github.com/jiseshen/HyperTrace}.
\end{abstract}

\section{Introduction}

Personalization is essential for large language models to function as effective partners in human-AI collaboration \citep{tseng2024two, liu2025survey, zhang2025personalization, xie2025a, guan2025survey}. Users differ in goals, preferences, and expectations, and these differences emerge gradually over repeated interactions. Effective personalization must therefore adapt continually while remaining scalable across diverse users.

Existing approaches can broadly fall into two categories. Training-based methods achieve strong personalization by learned user representations \cite{qiu-etal-2025-latent, ning2025user, liu2025llms+}, PEFT \cite{zhang2024personalized, tan2024democratizing}, or reinforcement learning \cite{jang2024personalized, poddar2024personalizing, jin2025erarealworldhumaninteraction, liang2026learning}, but incur computational overhead and depend on white-box models, limiting their applicability. Inference-level methods, in contrast, avoid parameter updates and instead condition generation on summary of history interactions \cite{wang2023cue, garbacea2025hyperaligninterpretablepersonalizedllm} or memory mechanisms \cite{salemi-etal-2024-lamp, zhong2024memorybank, zhang2024guided}. While more scalable, they typically represent user information as unstructured text or retrieved examples, offering limited interpretability and weak generalization beyond observed interactions.

\begin{figure}
    \centering
    \includegraphics[width=1.00\linewidth, trim=8pt 0pt 6pt 0pt, clip]{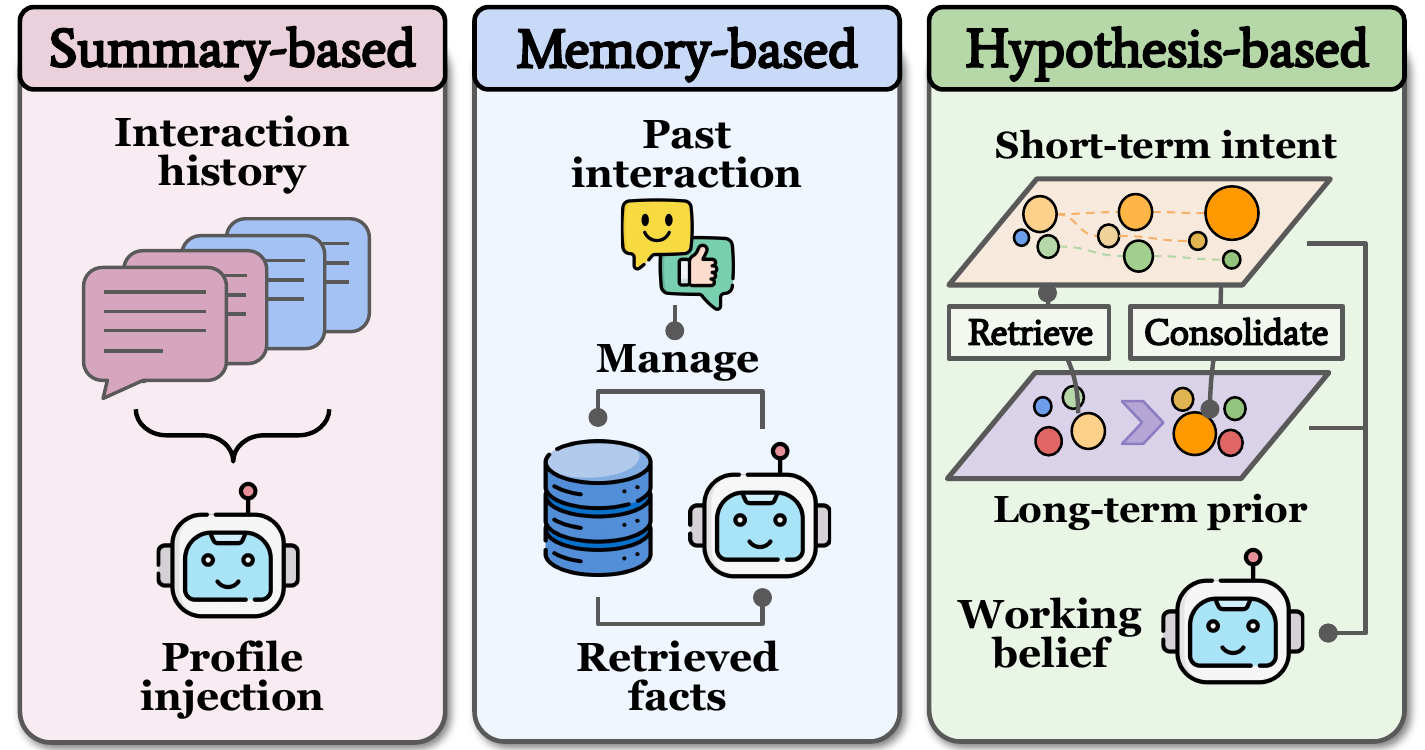}
    \caption{
\textbf{From storing user history to tracing user preferences.}
Existing training-free personalization methods compress interactions into a profile (left) or retrieve relevant memories (middle).
\methodname instead maintains interpretable hypotheses over short-term intent and long-term preferences (right), enabling continual and interpretable adaptation from user feedback.
}
    \vspace{-4.5mm}
    \label{fig:head}
\end{figure}

We instead formulate personalization as latent inference inspired by theory of mind \cite{kim2025hypothesisdriventheoryofmindreasoninglarge}. We treat user interactions as observations of latent variables that capture goals, preferences, and expectations of model behavior. A user profile is thus represented by several hypotheses about underlying intent, rather than adapted parameters or stored dialogue traces.
To handle sequential, sparse, and noisy signals, we use a Sequential Monte Carlo (SMC)-style procedure over natural-language hypotheses. Each particle represents a candidate interpretation of user intent, and feedback changes its weight under an LLM-based surrogate model of the observed choice. The weighted particle set preserves several competing explanations as evidence accumulates. We further organize hypotheses hierarchically: lower-level components capture short-term, topic-specific intent for rapid adaptation, while higher-level components consolidate stable long-term preferences.
Compared to training-based personalization, our method avoids parameter updates and extends to black-box endpoints. Compared to memory-based approaches, it preserves multiple interpretable preference explanations with trackable evidence accumulation. During generation, a summarized user profile conditions the model to produce user-specific responses.

We evaluate our framework in an online personalization setting based on PRISM~\citep{kirk2024prism} and PersonaMem-v2~\citep{personamemv2}. The evaluation covers response alignment, preference prediction, and profile alignment, directly measuring how well the extracted profile indicates user choice, and whether its adapted outputs move toward user-specific preferences. Results show stronger and more robust performance over summary- and memory-based baselines. Our contributions are thus twofold:

\begin{itemize}[leftmargin=*, itemsep=0pt, topsep=2pt, parsep=0pt, partopsep=0pt]
    \item We introduce a relative response-alignment score and build an online personalization evaluation framework from pluralistic alignment datasets, covering response-level adaptation, preference prediction, and profile consistency.
    \item We propose a training-free latent-preference inference framework that uses SMC-style importance weighting under a surrogate choice model, maintaining interpretable short- and long-term preference hypotheses without parameter update.
\end{itemize}

\section{Related Work}


\subsection{Evaluating LLM Personalization}

Personalization benchmarks extend LLM evaluation beyond generic instruction following by requiring models to condition on user-specific histories, preferences, or interaction traces. Existing benchmarks cover user-conditioned classification, retrieval, recommendation, generation, long-term conversational memory, dynamic profiles, implicit preferences, and multi-session interaction histories \citep{salemi-etal-2024-lamp, zollo2025personalllm, wu2025longmemeval, personamemv2, zhao2025do, jiang2025know, jin2026thoughttrace}. Preference-feedback datasets further expose heterogeneous human preferences through user-specific choices or in-situ feedback over candidate responses \citep{kirk2024prism, castricato2025persona, shi2024wildfeedback}.

These benchmarks provide complementary evaluation signals, including task labels, selected candidates, reference responses, and reward-model scores \citep{salemi-etal-2024-lamp, personamemv2, dong-etal-2024-llm, zollo2025personalllm}. Our evaluation builds on this line by converting pluralistic preference data into an online personalization setting and measuring response alignment, preference prediction, and profile alignment.

\subsection{Methods for LLM Personalization}

One family of methods personalizes LLMs through training-based adaptation. These approaches learn user representations \citep{qiu-etal-2025-latent, ning2025user, liu2025llms+}, PEFT modules \citep{zhang2024personalized, tan2024democratizing}, or personalized post-training objectives based on reward modeling, RLHF, or user feedback \citep{jin2025erarealworldhumaninteraction, jang2024personalized, poddar2024personalizing, liang2026learning}. Related work also studies learned memory construction, group-level personalization, black-box-compatible external modules, and amortized or meta-learning mechanisms \citep{magister2024wayllmpersonalizationlearning, zhang2025proper, zhuang2024hydra, tan2025instant, personamemv2}.

Another family performs personalization at inference time without updating the base model. Existing approaches summarize user histories into natural-language profiles or personalized prompts \citep{zhang2024guided, richardson2023integrating, li2024learning, qiu2025measuring, wang2023cue, garbacea2025hyperaligninterpretablepersonalizedllm}, retrieve relevant user histories or optimize retrieved contexts for personalized generation \citep{salemi-etal-2024-lamp, salemi2024optimization}, and maintain persistent user records or long-term memories across sessions \citep{zhong2024memorybank, madaan2022memory, dalvi2022towards}. Decoding-time steering provides another efficient route, but requires control over the sampling process \citep{chen2025pad, zhang2025amulet}.

\methodname is closest to inference-time personalization, but represents user information as a weighted set of natural-language hypotheses rather than a single profile, retrieved context, or memory state. Its update is related to Bayesian hypothesis filtering \citep{kim2025hypothesisdriventheoryofmindreasoninglarge}, but focuses on tracing
actionable preference signals from user choices.

\begin{figure*}[t]
    \centering
    \includegraphics[width=1\linewidth, trim=9pt 0pt 9pt 0pt, clip]{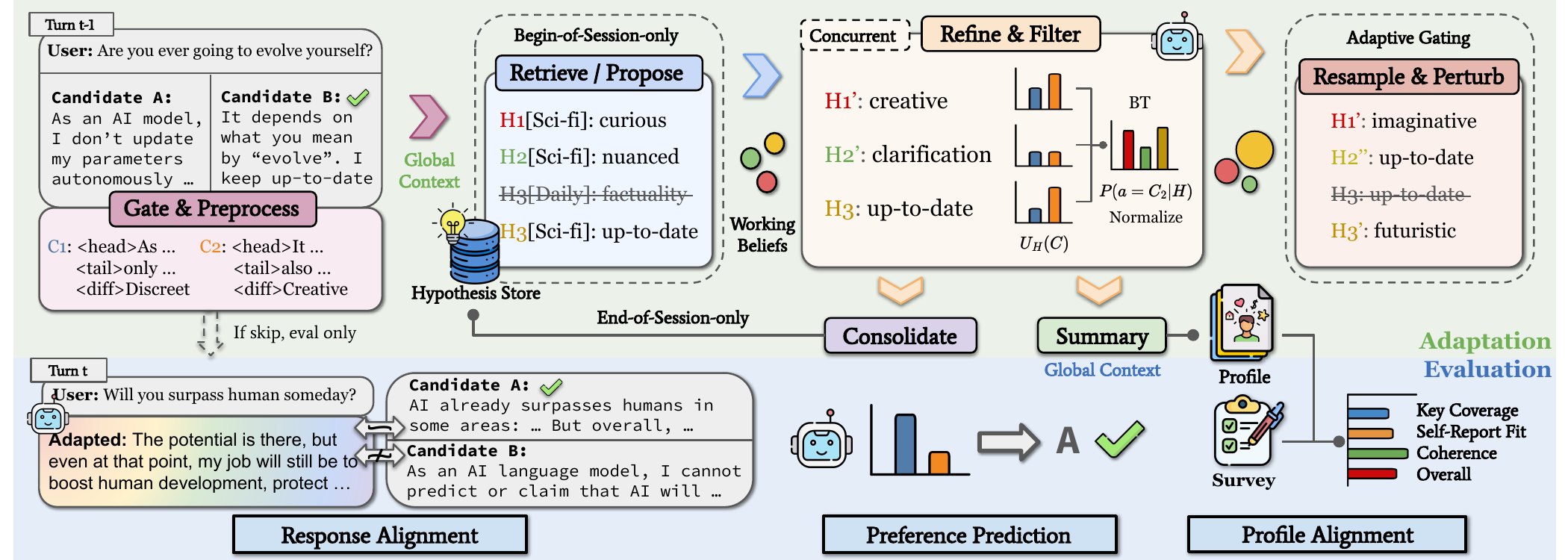}
    \caption{
\textbf{Overview of the intertwined adaptation--evaluation loop of \method.} During adaptation, preference-bearing turns are gated and compressed into global contexts for retrieving or proposing (if inapplicable) session-level hypotheses. 
Working beliefs are refined and filtered by estimating hypothesis-conditioned utilities and converting them into surrogate Bradley--Terry choice scores. 
Adaptive particle maintenance triggers resampling via the effective sample size and perturbs detected similarity groups, preventing belief collapse. 
At session end, beliefs are consolidated into long-term memory and summarized as a profile. During evaluation, the profile serves as global context, and held-out interactions measure response alignment, preference prediction, and profile alignment.
}
    \label{fig:main}
\end{figure*}

\section{Preliminary}

We model personalized chatbot interaction as a sequence of sessions in which the active user preference is latent and context-dependent. At the beginning of a session $s$, the user's current need $r_s$ is sampled from their broader interest space, inducing a session-specific preference state $z_s$:
\begin{equation}
    r_s \sim p(r\mid u), \qquad z_s \sim p(z\mid u,r_s).
\end{equation}
This captures the intuition that the same user may prefer different response styles across intents. For example, code-related queries favor direct and actionable answers, whereas open-ended daily questions favor broader and more engaging responses.

Within a session, the user reveals preferences through interaction feedback. Given a query $x_t$ and a candidate set $C_t=\{c_{t,1},\ldots,c_{t,m}\}$, a Bradley--Terry model provides a convenient observation model for the user's selection:
\begin{equation}
\label{eq:bradley-terry}
P(a_t=i\mid x_t,C_t,z_s)
=
\frac{\exp(U_{x_t,z_s}(c_{t,i}))}
{\sum_j \exp(U_{x_t,z_s}(c_{t,j}))}.
\end{equation}
Here, $U_{x_t,z_s}(c)$ denotes the utility of candidate $c$ under the current query and latent preference state. If this utility and the prior over $z_s$ were known, the posterior preference belief would satisfy
\begin{equation}
p(z_s\mid \mathcal{D}_{1:t})
\propto
p(z_s)\prod_{\ell=1}^{t}
P(a_\ell\mid x_\ell,C_\ell,z_s),
\end{equation}
where $\mathcal{D}_{1:t}$ denotes the observed feedback history.

\section{Methodology}

\begin{table}[t]
\centering
\small
\setlength{\tabcolsep}{4pt}
\renewcommand{\arraystretch}{1.5}
\caption{Notation used in \methodname formulation.}
\vspace{-1mm}
\begin{tabularx}{\columnwidth}{@{}lX@{}}
\toprule
\textbf{Symbol} & \textbf{Meaning} \\
\midrule
$r_s, z_s$ & Need and latent preference state behind session $s$ \\
$x_t, C_t, a_t$ & Query, candidate responses, and observed selection at turn $t$ \\
$\mathcal{D}_{1:t}$ & Interaction feedback observed through turn $t$ \\
$h_t^{(k)}, w_t^{(k)}$ & Hypothesis $k$ and its normalized relative support \\
$\mathcal{B}_t$ & Five-hypothesis working belief at turn $t$ \\
$\mathcal{M}_u$ & Persistent hypothesis memory for user $u$ \\
$\hat U, \hat P$ & LLM-estimated utility and choice probability \\
$\mathcal{T}_s, H(\mathbf w)$ & Valid traced turns and normalized belief entropy \\
$\mathrm{RA}(y_t)$ & Relative response alignment of adapted output $y_t$ \\
\bottomrule
\end{tabularx}
\label{tab:notation}
\vspace{-2mm}
\end{table}

\subsection{HyperTrace}

\methodname represents the possible preference state $z_s$ with a set of $K$ weighted natural-language hypotheses. Each hypothesis describes a possible explanation of the user's current preference, such as preferring concise implementation details, broader conceptual explanations, or more cautious wording. At turn $t$, the working belief is
\begin{equation}
    \mathcal{B}_t=\{(h_t^{(k)},w_t^{(k)})\}_{k=1}^{K}.
\end{equation}

The normalized weights rank the maintained explanations according to how well their induced choice scores account for the observed feedback, while retaining plausible alternatives.

\subsubsection{Intra-Session Belief Update}
\label{sec:intra_session_updating}

As illustrated in \autoref{fig:main}, belief update consists of four steps. \textit{Refine} adapts the hypotheses to new evidence. \textit{Filter} scores each hypothesis against the observed feedback. \textit{Resample} retains hypotheses with greater relative support. \textit{Perturb} introduces distinct preference axes to avoid premature collapse.

In each filtering step, \methodname updates hypothesis weights according to how well each hypothesis predicts the user's observed selection. For each hypothesis $h_t^{(k)}$, we prompt the LLM to approximate the user's utility function by scoring each candidate response:
\begin{equation}
\hat{U}_{x_t,h_t^{(k)}}(c_{t,i})
=
f_{\theta}(x_t, c_{t,i}, h_t^{(k)}).
\end{equation}
We convert these hypothesis-conditioned utilities into a surrogate choice score using Eq.~\ref{eq:bradley-terry}. The particle weights are updated by
\begin{equation}
w_t^{(k)}
\propto
w_{t-1}^{(k)}
\hat{P}(a_t \mid x_t,C_t,h_t^{(k)}).
\end{equation}
Thus, the LLM supplies the hypothesis-conditioned utility estimate, while the fixed choice rule converts chosen-versus-rejected contrasts into comparable importance weights.

The five-slot lifecycle follows this update throughout a session. At the first usable turn, the initial message retrieves five topic-aware memory items, and the initializer either reuses them or proposes new hypotheses slot by slot. At later usable turns, each slot is independently revised or replaced before the surrogate scores reweight the five hypotheses. We resample when the effective sample size falls below the threshold, then group exact duplicates and hypotheses with embedding similarity at least $\tau=0.8$. Each non-singleton group $G$ is merged into one hypothesis, and the remaining $|G|-1$ slots are repopulated along preference axes not yet represented. At session end, the final five-slot belief is consolidated into $\mathcal{M}_u$.

We also include a lightweight gate before tracing. Many real interactions, such as greetings, typos, or clarification turns, do not provide reliable preference evidence. The gate skips such turns and carries the belief state forward unchanged. To reduce context cost, we preprocess long candidate responses into compact summaries while preserving their most preference-relevant content.

Finally, the tracing procedure is naturally parallelizable \cite{kim2025hypothesisdriventheoryofmindreasoninglarge}. Branching and filtering over different hypotheses can be executed independently, and their shared prompt prefixes allow substantial KV cache reuse, keeping inference overhead manageable. We further use a routing routine that assigns different model backends to substeps according to their complexity. As shown in \S\ref{sec:cost_quality_tradeoff}, this model routing strategy further reduces cost without degrading tracing quality.

\subsubsection{Cross-Session Preference Consolidation}

Intra-session beliefs capture the user's active preference in the current conversation, but personalization also requires stable memory across sessions. We therefore maintain a persistent hypothesis set $\mathcal{M}_u$ for each user. At a new session, the initial message retrieves the five nearest hypotheses using embeddings of both topic labels and hypothesis content. The structured initializer then decides, slot by slot, whether to reuse a retrieved item or create a new hypothesis. The same procedure is restarted after a detected major topic shift.

At the end of a session, the final working beliefs are smoothed back into memory according to
\begin{equation}
p'_{\mathcal{M}}(h_i)
=
(1-\alpha_s)p_{\mathcal{M}}(h_i)
+
\alpha_s w_i,
\end{equation}
We define the consolidation weight of the session as
\begin{equation}
\alpha_s =
(1-e^{-|\mathcal{T}_s|})\sqrt{1-H(\mathbf{w})},
\end{equation}
where $|\mathcal{T}_s|$ is the number of valid traced turns and $H(\mathbf{w})$ is the normalized entropy of the final belief weights. Thus, $\alpha_s$ controls how strongly the current session contributes to memory based on the amount and decisiveness of its evidence. It does not determine where a preference should apply: topical scope is handled separately by topic-labelled storage and context-conditioned retrieval.

Each stored hypothesis is associated with topic metadata which can be used in later retrieval. When a hypothesis repeatedly explains user behavior across different topics, its scope gradually broadens; when it only applies within a narrow context, it remains topic-specific. This design helps distinguish stable cross-topic preferences from temporary task-specific needs. If a major topic shift is detected within a conversation, we treat the subsequent turns as a new session and restart retrieval.

\subsection{Online Personalization Evaluation}

\subsubsection{Evaluation Framework}

We evaluate online personalization from three complementary perspectives: response alignment, preference prediction, and profile alignment.

\noindent \textbf{Response Alignment.}
Given an adapted response $y_t$, the user's chosen candidate $c_t^+$, and rejected candidates $C_t^- = C_t \setminus \{c_t^+\}$, we define:
\begin{equation}
\mathrm{RA}(y_t)
=
S(y_t,c_t^+)
-
\max_{c\in C_t^-}S(y_t,c).
\end{equation}
Here, $S$ is instantiated either as embedding cosine similarity or as an LLM-based similarity score. We use a relative score because the chosen candidate is not an absolute gold response, but the user's preferred option among the displayed candidates. Thus, RA measures whether personalization moves the adapted response closer to the user's selected response than to rejected alternatives.

\noindent \textbf{Preference Prediction.}
We ask the personalized agent to predict which candidate response the user will choose given the interaction history and the inferred user profile. This evaluation is not intended to benchmark the underlying LLM's raw prediction ability. Instead, it tests whether the generated profile contains the preference-relevant information needed to recover the user's observed choices.

\noindent \textbf{Profile Alignment.}
We ask each method to summarize the user's profile, comparing it against user-side evidence, such as survey responses or system prompts. Since generated profiles and user-written evidence may differ substantially in surface form, exact matching is inappropriate. We therefore use rubric-based LLM evaluation over key-aspect coverage, contradiction avoidance, specificity, and overall consistency. This metric evaluates whether the inferred long-term memory semantically aligns with explicit user evidence, beyond being behaviorally useful for prediction or generation.

For all LLM-based evaluation, we use a discrete 0--5 rubric with explicit criteria and examples \citep{0-5scale}. We use \texttt{gemini-3-flash} as the default evaluator for response alignment, preference prediction, and profile alignment. To reduce self-preference bias, inference and evaluation are performed by models from different providers \citep{self-preference}. We additionally repeat the evaluation with \texttt{claude-sonnet-4.6} under the same protocol (\S\ref{sec:robustness_analyses}). We also report embedding-based response alignment as a complementary automatic metric, using its trend-level agreement with LLM-based scores as convergent evidence. For embedding-based similarity, we use \texttt{text-embedding-3-small}.

\begin{table}[t]
\centering
\caption{
Dataset sampling summary for PRISM~\cite{kirk2024prism} and PersonaMem-v2~\cite{personamemv2}. We first filter users with over 20 turns, then randomly sample 50 eligible users per dataset for evaluation.
}
\setlength{\tabcolsep}{3.8pt}
\renewcommand{\arraystretch}{1.08}
\resizebox{\columnwidth}{!}{

\begin{tabular}{lccccc}
\toprule
\textbf{Dataset / split} 
& \multicolumn{2}{c}{\textbf{Source}} 
& \multicolumn{3}{c}{\textbf{Sampled}} \\
\cmidrule(lr){2-3}\cmidrule(lr){4-6}
& \textbf{Users} & \textbf{Turns} 
& \textbf{Eligible} & \textbf{Users} & \textbf{Turns} \\
\midrule
PRISM
& 1,396
& 27,170
& 621 
& 50
& 1,206 \\
PersonaMem-v2 (test)
& 200 
& 5,000
& 162 
& 50
& 1,335 \\
\bottomrule
\end{tabular}
}
\label{tab:dataset_sampling}
\vspace{-3mm}
\end{table}

\begin{table*}[h]
\centering
\small
\setlength{\tabcolsep}{3.2pt}
\renewcommand{\arraystretch}{1.12}
\caption{
Main results on PRISM and PersonaMem-v2.
Acc$_{>20}$ and $\Delta$Acc measure preference-prediction accuracy and its gain over turn 0.
GPT$_{>20}$ and Emb$_{>20}$ measure response alignment after 20 interactions using LLM-judge and embedding-based relative similarity, respectively, with $\Delta$ columns denoting gains over turn 0.
Prof. and Sim. measure final profile alignment: Prof. is a rubric-based LLM profile score, and Sim. is embedding similarity to ground-truth survey.
All online $>20$ metrics average turns after turn 20 with at least 10 active users.
}
\begin{tabular*}{\textwidth}{@{\extracolsep{\fill}}llcccccccc@{}}
\toprule
\multirow{2}{*}{\textbf{Dataset}} 
& \multirow{2}{*}{\textbf{Method}}
& \multicolumn{2}{c}{\textbf{Preference Prediction}}
& \multicolumn{4}{c}{\textbf{Response Alignment}}
& \multicolumn{2}{c}{\textbf{Profile Alignment}} \\
\cmidrule(lr){3-4} \cmidrule(lr){5-8} \cmidrule(lr){9-10}
& 
& Acc$_{>20}$ 
& $\Delta$Acc 
& GPT$_{>20}$ 
& $\Delta$GPT 
& Emb$_{>20}$ 
& $\Delta$Emb 
& Prof. 
& Sim. \\
\midrule

\multirow{6}{*}{PRISM}
& CoT
& 0.5157
& 0.0728
& -0.2223
& 0.2563
& -0.0302
& 0.0049
& --
& -- \\

& RAG
& 0.4694
& 0.0765
& -0.3400
& 0.2600
& -0.0260
& \textbf{0.0396}
& --
& -- \\

& Cheatsheet
& \underline{0.5436}
& 0.1293
& -0.1433
& \textbf{0.4210}
& -0.0124
& \underline{0.0365}
& 3.9143
& \underline{0.4952} \\

& HyperAlign
& 0.4959
& 0.0399
& \underline{-0.0149}
& 0.0971
& \underline{0.0042}
& 0.0155
& \underline{3.9429}
& 0.4924 \\

& Hydra
& 0.5080
& \underline{0.2151}
& --
& --
& --
& --
& --
& -- \\

\cmidrule(lr){2-10}
& HT (ours)
& \textbf{0.6136}
& \textbf{0.2207}
& \textbf{0.0739}
& \underline{0.3953}
& \textbf{0.0093}
& 0.0334
& \textbf{4.1857}
& \textbf{0.5366} \\

\midrule

\multirow{6}{*}{PersonaMemV2}
& CoT
& 0.2922
& -0.0006
& -0.5393
& \underline{0.2036}
& -0.0391
& 0.0012
& --
& -- \\

& RAG
& 0.2793
& -0.0850
& -0.5122
& 0.0450
& \underline{-0.0360}
& \underline{0.0026}
& --
& -- \\

& Cheatsheet
& \underline{0.3035}
& \underline{0.0249}
& \underline{-0.4704}
& \textbf{0.3072}
& \textbf{-0.0325}
& \textbf{0.0059}
& 4.2500
& 0.6389 \\

& HyperAlign
& 0.2895
& -0.0462
& -0.5235
& 0.0382
& -0.0368
& -0.0096
& \textbf{4.3821}
& \textbf{0.6728} \\

& Hydra
& 0.2617
& 0.0046
& --
& --
& --
& --
& --
& -- \\

\cmidrule(lr){2-10}
& HT (ours)
& \textbf{0.3768}
& \textbf{0.1554}
& \textbf{-0.4388}
& 0.1112
& -0.0365
& 0.0002
& \underline{4.3589}
& \underline{0.6612} \\

\bottomrule
\end{tabular*}
\label{tab:main_results}
\end{table*}

\begin{figure*}[h]
    \centering
    \includegraphics[width=1\linewidth, trim=5pt 10pt 10pt 10pt, clip]{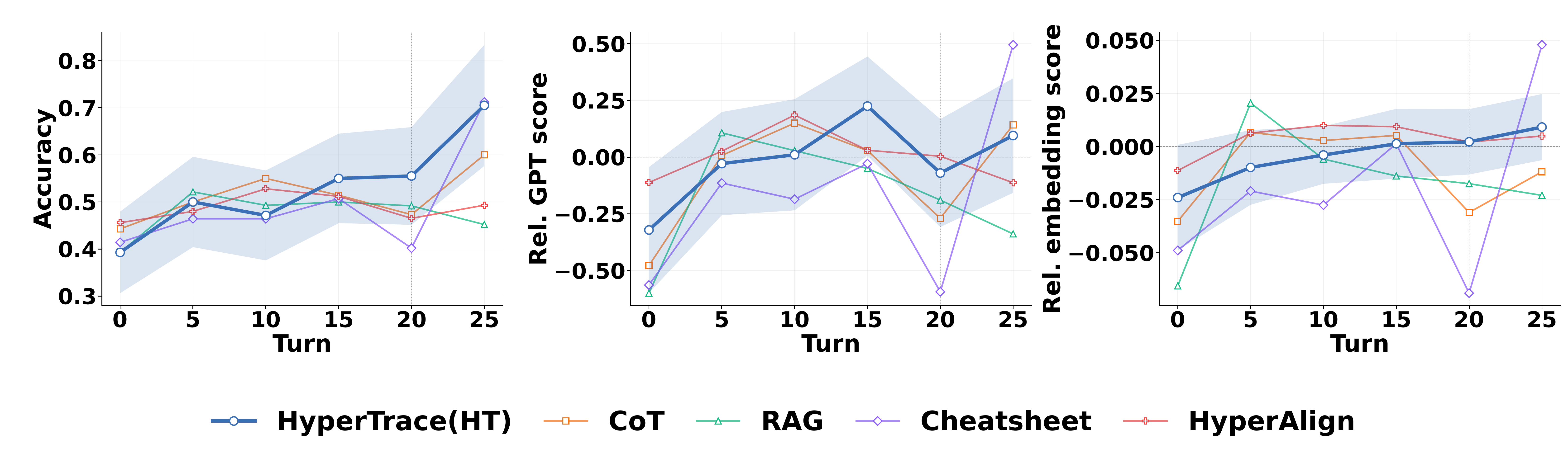}
    \caption{
Overall comparison on PRISM. Shaded bands denote 95\% user-level bootstrap confidence intervals, and the dotted line marks the main session-transition region. \methodname remains more stable across this transition and improves by the final turn.
}
    \label{fig:main_comparison}
    \vspace{-3mm}
\end{figure*}

\section{Experiments}

\subsection{Main Comparison}

\noindent \textbf{Experimental setting.}
Unless otherwise specified, \methodname uses \texttt{gpt-5} as the tracing model and maintains a working belief of $K=5$ preference hypotheses per user. We use \texttt{text-embedding-3-small} for embedding-based retrieval. All LLM baselines use the same \texttt{gpt-5} backbone with the same guidance prompts regarding preference extraction and response generation applied. For clarity, we summarize most online evaluation curves using two statistics: the average performance after 20 adaptation turns and the improvement relative to the first turn. Appendix~\ref{app:implementation-details} further details the implementation, and Appendix~\ref{app:prompt-family} documents prompts and dataset adapters.

\noindent \textbf{Datasets.}
We evaluate \methodname and baselines on two personalization and alignment datasets: PRISM~\citep{kirk2024prism} and the test split of PersonaMem-v2 ~\citep{personamemv2}. PRISM contains real-world feedback covering participants born in 75 countries and residing in 38 countries, with survey data about communication preference. PersonaMem-v2 provides simulated personas with multi-turn feedback data, and additionally tested confounded negative preferences, and sensitive memory boundaries. Since our setting requires sufficient cross-turn evidence for online preference inference, we filter users with more than 20 total turns and sample 50 eligible users from each dataset for evaluation, as summarized in \autoref{tab:dataset_sampling}.

\noindent \textbf{Baselines.}
We adapt all baselines to an online setting strictly conditioned on past interactions at each turn. 
CoT \cite{cot} uses the latest 5 observed turns as few-shot examples and internally reasons about the user preference. 
RAG \cite{rag} instead retrieves 5 semantically similar past interactions as examples. 
Dynamic Cheatsheet \cite{suzgun2025_DynamicCheatsheet} incrementally updates a compact user-preference summary after each observed choice, capped at 10 bullet points. 
HyperAlign \cite{garbacea2025hyperaligninterpretablepersonalizedllm} extracts 5 relevant hypotheses from the whole interaction history for each prediction. 
Hydra \cite{zhuang2024hydra} acts as a \texttt{RoBERTa}-based supervised reranker trained on 1,000 out-of-bag turns with a 5-turn history window; because it only ranks candidates, we exclude it from adapted-response generation results.

\noindent \textbf{\methodname achieves stable online adaptation.}
As shown in \autoref{tab:main_results}, \methodname leads preference prediction and response alignment on both datasets. On PRISM it also gives the strongest final profile alignment. On PersonaMem-v2, its profile scores do not exceed HyperAlign but are competitive. Therefore, the results support behavioral adaptation more clearly than a uniform advantage in profile reconstruction.
\autoref{fig:main_comparison} further shows that \methodname maintains a relatively stable adaptation trajectory, especially around the session-transition region between turns 18--22, where 41/50 users move to a new session. CoT and RAG obtain stronger early response-alignment scores because they directly access candidate responses as reference, but their gains are not sustained in later turns. Dynamic Cheatsheet adapts quickly to new topics and preferences, yet its sharp drop at session changes suggests weaker cross-session stability. HyperAlign remains competitive, but its capacity becomes bounded under sufficiently long contexts. Overall, these gains may reflect our method’s targeted design for maintaining fine-grained preference hypotheses while consolidating persistent signal across sessions.
Detailed plots for both PRISM and PersonaMem-v2 are retained in
Appendix~\ref{app:detailed-turn-level-results}.

Absolute response alignment remains negative for all methods on PersonaMem-v2, where each choice is compared against multiple rejected candidates. Under the shared protocol, the scores remain meaningful for relative comparison, while the benchmark’s boundary cases motivate safety-aware filtering of hypotheses as a potential extension.

\begin{table}[t]
\centering
\caption{
Structural ablations on PRISM. Acc$_{>20}$ and GPT$_{>20}$ are post-turn-20 averages; Prof. and Cost denote profile alignment and USD per tracing turn. The hybrid deployment configuration is reported separately in \S\ref{sec:cost_quality_tradeoff}.
}
\small
\setlength{\tabcolsep}{3.8pt}
\renewcommand{\arraystretch}{1.08}
\begin{tabular*}{\columnwidth}{@{\extracolsep{\fill}}lcccc@{}}
\toprule
\textbf{Method} & Acc$_{>20}$ & GPT$_{>20}$ & Prof. & Cost $\downarrow$ \\
\midrule
HT (GPT-5)
& \textbf{0.6136}
& \underline{0.0739}
& \underline{4.1857}
& 0.028010 \\
No-gating
& 0.5054
& 0.0944
& \textbf{4.2500}
& 0.040998 \\
Flat-5
& 0.5489
& 0.0313
& 4.1714
& 0.025396 \\
No-topic
& 0.5567
& -0.0500
& 4.1214
& \textbf{0.020916} \\
No-consol
& 0.5459
& \textbf{0.1100}
& 4.0857
& 0.025291 \\
\bottomrule
\end{tabular*}

\label{tab:method_ablation}
\vspace{-3mm}
\end{table}

\subsection{Method Ablation}

\noindent \textbf{Experimental setting.} \autoref{tab:method_ablation} keeps the dataset subset, evaluator, prompts, and hypothesis budget fixed, and changes one structural component at a time. We ablate skip gating, hierarchical memory, topic grounding, and consolidation from the full \texttt{gpt-5} tracer. The Flat-5 variant keeps only a five-hypothesis working belief without a persistent hypothesis store, while the other variants preserve the overall tracing pipeline and remove only the targeted mechanism. The routed hybrid is treated separately as an efficiency configuration in \S\ref{sec:cost_quality_tradeoff}, rather than as a structural ablation.

\noindent \textbf{Not every interaction is preference evidence.}
Removing gating substantially increases cost while reducing preference-prediction accuracy, indicating that many turns provide no useful preference evidence. In PRISM, 49.29\% of turns are skipped and not sent through the tracing update. These skipped turns often correspond to greetings and generic quality gaps weakly related to stable user preferences. Filtering them prevents noisy evidence from being absorbed into the hypothesis state and makes the method more suitable for realistic user interactions, where preference feedback is sparse. At the same time, No-gating still obtains a strong final profile score, suggesting that retrieval provides some robustness against noisy updates.

\noindent \textbf{Hierarchical hypotheses stabilize long-term preference tracking.}
These ablations highlight the role of the hierarchical belief structure. Compared with Flat-5, the full hypothesis store expands the global representational space beyond 5 online hypotheses, allowing the system to preserve preferences from multiple topics and interaction phases. The weaker No-topic result suggests that topic metadata is useful for both interpreting preference evidence and retrieving relevant hypotheses, consistent with the intuition that user preferences are often topic-dependent. Finally, No-consol shows that directly synchronizing short-term working beliefs back into the store is less reliable than hierarchical consolidation, which buffers transient topic fluctuations and stabilizes long-term preferences.

\subsection{Robustness Analyses}
\label{sec:robustness_analyses}

\begin{figure}[t]
    \centering
    \begin{subfigure}[t]{0.49\linewidth}
        \includegraphics[width=\linewidth]{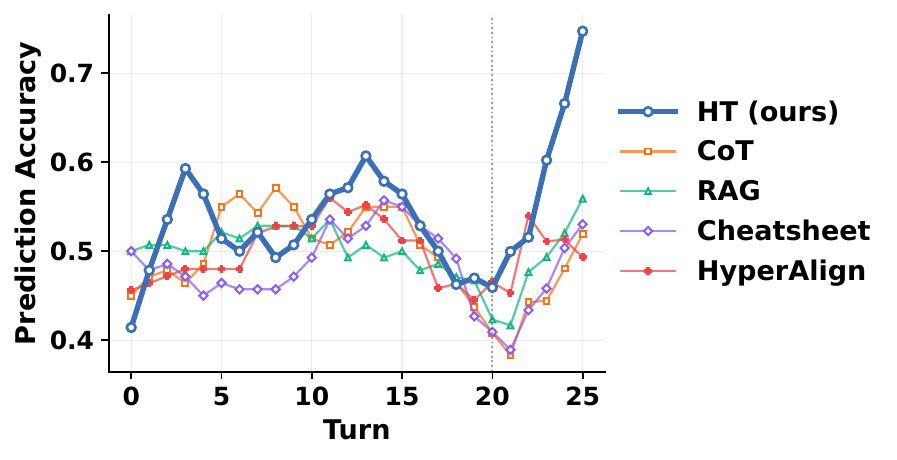}
        \caption{preference prediction.}
    \end{subfigure}
    \hfill
    \begin{subfigure}[t]{0.49\linewidth}
        \includegraphics[width=\linewidth]{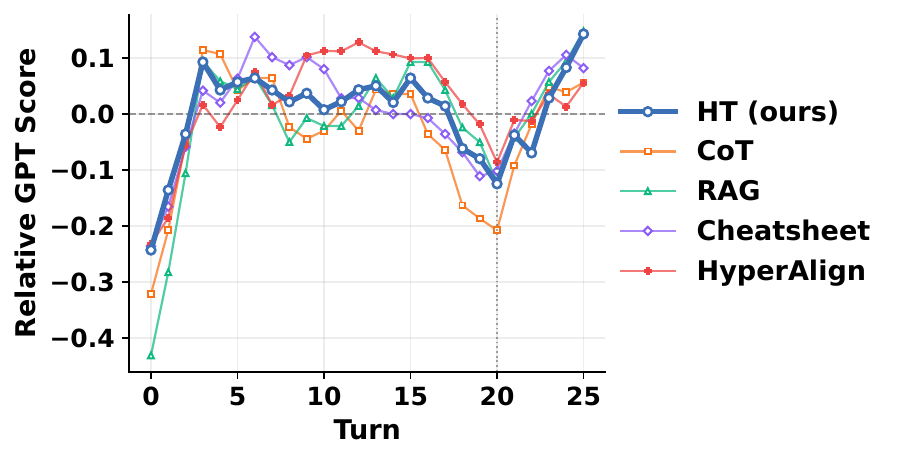}
        \caption{Response alignment.}
    \end{subfigure}

    \vspace{1mm}

    \begin{subfigure}[t]{0.55\linewidth}
        \includegraphics[width=\linewidth]{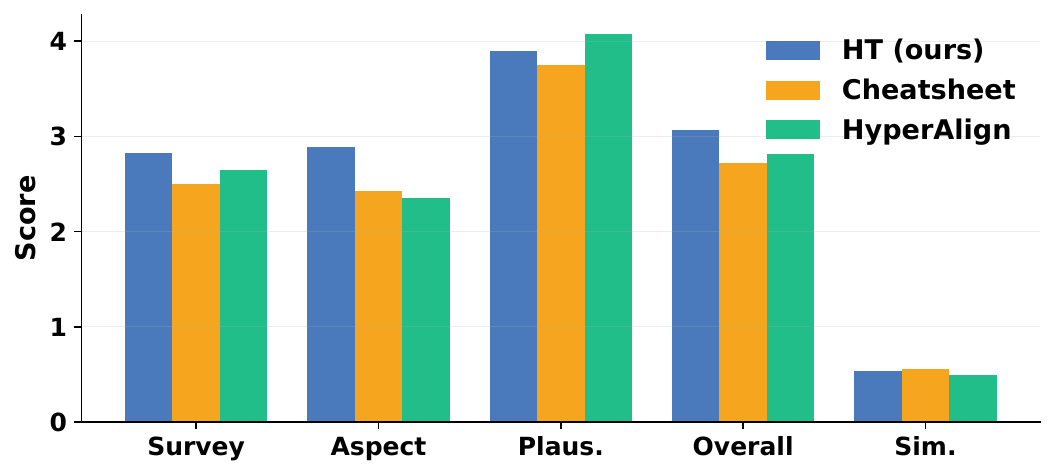}
        \caption{Profile alignment.}
    \end{subfigure}
    \caption{Cross-evaluator robustness.}
    \label{fig:claude_eval_results}
\end{figure}

\begin{figure}[t]
    \centering
    \includegraphics[width=0.94\linewidth, trim=0pt 15pt 0pt 10pt, clip]{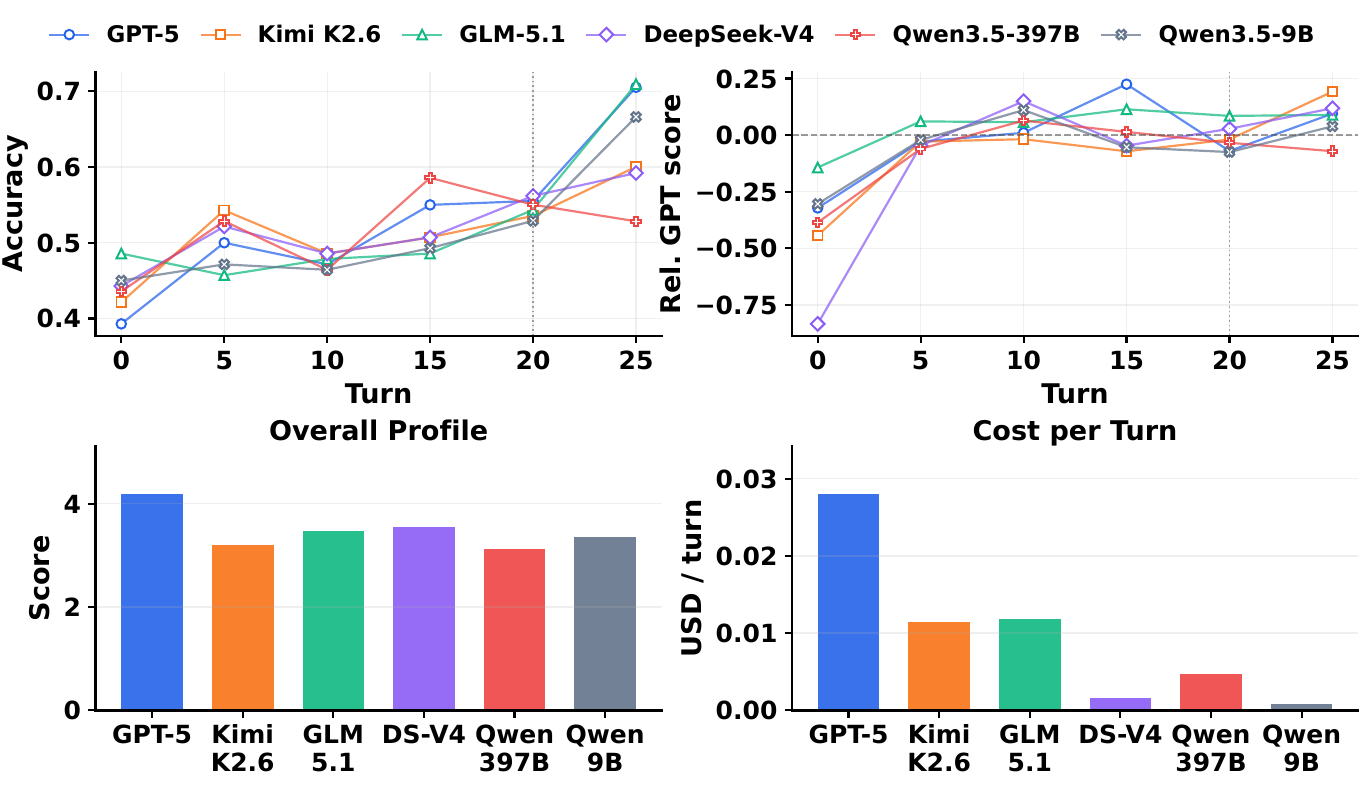}
    \caption{Tracing-backbone sensitivity on PRISM, including adaptation curves, profile alignment, and cost.}
    \label{fig:model_ablation}
    \vspace{-3mm}
\end{figure}

\begin{table}[t]
\centering
\footnotesize
\renewcommand{\arraystretch}{1.2}
\setlength{\tabcolsep}{2pt}
\caption{
Robustness of \methodname on PRISM under sparse feedback and
frequent topic shifts.
}
\begin{tabular*}{\columnwidth}{
    @{\extracolsep{\fill}}llccc@{}
}
\toprule
\textbf{Stress test}
& \textbf{Method}
& Acc$_{>20}$
& GPT$_{>20}$
& Prof. \\
\midrule

\multirow{2}{*}{Full feedback}
& HT
& \textbf{0.5878}
& \textbf{0.1219}
& \textbf{4.2500} \\
& Cheatsheet
& 0.5436
& -0.1433
& 3.9143 \\

\multirow{2}{*}{80\% withheld}
& HT
& \textbf{0.5567}
& \textbf{0.1002}
& \textbf{3.9333} \\
& Cheatsheet
& 0.4582
& -0.0232
& 3.2000 \\

\multirow{2}{*}{Frequent shifts}
& HT
& \textbf{0.5304}
& \textbf{0.0943}
& \textbf{4.1400} \\
& Cheatsheet
& 0.4224
& -0.0193
& 3.5300 \\

\bottomrule
\end{tabular*}
\label{tab:online_robustness}
\end{table}

\paragraph{Robustness to sparsity and frequent topic shifts.}
As shown in \autoref{tab:online_robustness}, \methodname assumes explicit comparative choices. To test sparse supervision, we withhold 80\% of the available choice updates; \methodname remains ahead of Dynamic Cheatsheet on all three measures. Separately, we select the 20 PRISM users with the most sessions—and therefore the shortest sessions on average—to test sensitivity to frequent topic transitions. On this cohort, \methodname improves preference-prediction accuracy within the first noninitial session from 0.3917 to 0.4717, suggesting that its topic-conditioned memory can preserve useful information while adapting across rapidly changing contexts.

\paragraph{The trends persist across evaluators and tracing backbones.}
Repeating the evaluation with \texttt{claude-sonnet-4.6} preserves the overall
ranking and trends (\autoref{fig:claude_eval_results}), suggesting that the
comparison is not specific to the default evaluator. We also run the tracing
procedure with five alternative backbones while keeping
\texttt{gemini-3-flash} as the judge. As \autoref{fig:model_ablation} shows,
all backbones improve preference-prediction accuracy from turn 0, and most
maintain near-positive or positive late-stage response alignment. Performance
is not monotonic in model size, but the overall adaptation pattern is largely
consistent, indicating that the tracing scaffold is not tied to one backbone.
Additional tests of initialization, particle propagation, and perturbation are
reported in Appendix~\ref{app:additional-tracing-analyses}.

\subsection{Cost--Quality Trade-off}
\label{sec:cost_quality_tradeoff}

\begin{figure}[t]
    \centering
    \includegraphics[width=1\linewidth, trim=13pt 0pt 5pt 0pt ,clip]{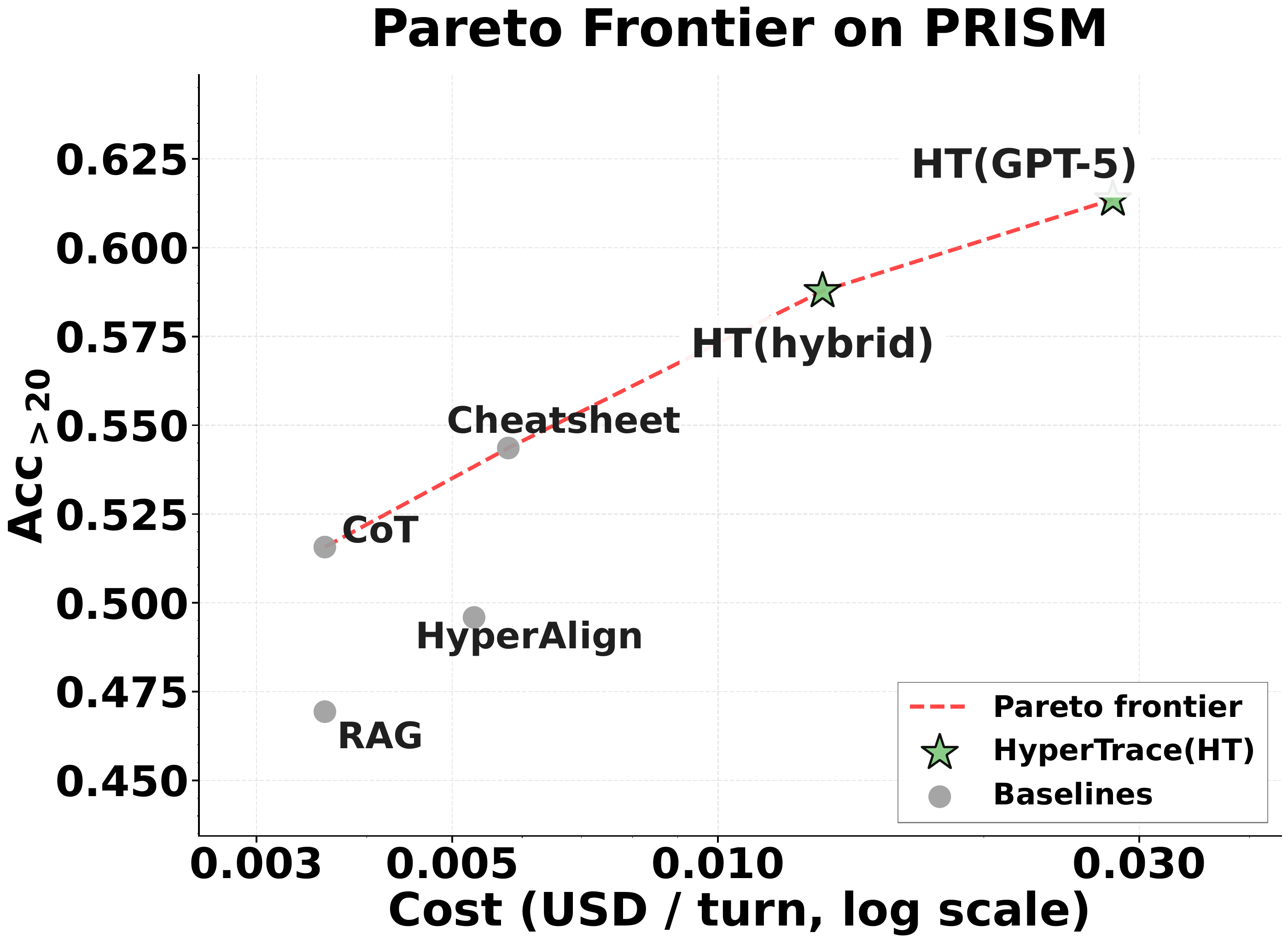}
    \caption{
    Cost--accuracy trade-off on PRISM, measured by online cost per turn and Acc$_{>20}$.
    HT(\texttt{gpt-5}) gives the strongest accuracy, while HT(hybrid) provides a more practical Pareto point.
    }
    \label{fig:prism_pareto_acc_cost}
    \vspace{-5mm}
\end{figure}

\paragraph{\methodname shifts the Pareto frontier.}
Beyond accuracy, online personalization must remain cost-effective because adaptation is performed repeatedly over user interactions.
\autoref{fig:prism_pareto_acc_cost} reports PRISM cost per online turn against Acc$_{>20}$.
HT(\texttt{gpt-5}) is the quality-oriented reference used in the main comparison and reaches the highest accuracy, 0.6136 Acc$_{>20}$. HT(hybrid) is a separate deployment configuration: it obtains 0.5878 Acc$_{>20}$ at \$0.0131 per turn, reducing cost by 53.1\% while retaining 95.8\% of the reference accuracy. It also has higher observed response alignment (0.1219 vs. 0.0739) and profile alignment (4.2500 vs. 4.1857). There is therefore no single configuration that dominates every criterion; the full model provides the cleanest like-for-like quality comparison, while hybrid routing is the more practical Pareto point.

\subsection{Qualitative Analysis}

\begin{figure}
    \centering
    \vspace{-2mm}\includegraphics[width=1\linewidth]{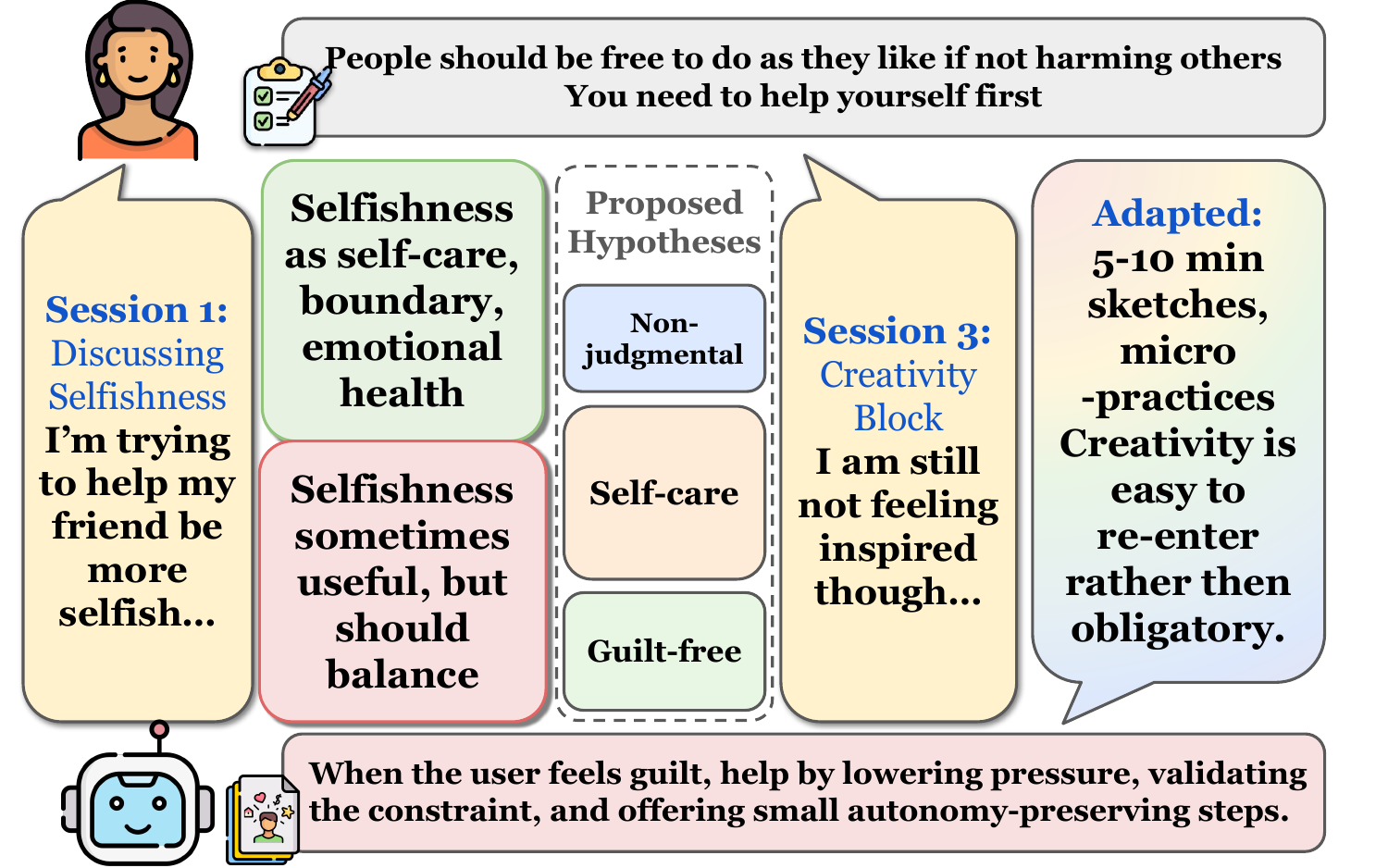}
    \caption{
An example of successful cross-session \methodname from user \texttt{516} in PRISM.}
    \label{fig:case_study}
    \vspace{-3mm}
\end{figure}

\paragraph{Transferable Preferences Beyond Topic Memory.}
In the case depicted in \autoref{fig:case_study}, a self-care preference learned from interpersonal advice is later retrieved and operationalized in a creative-habit setting: the system shifts from generic art-block advice to low-pressure, guilt-free micro-practices, showing that the learned preference is
stored as a transferable behavioral tendency rather than a topic-specific memory. Appendix~\ref{app:illustrative-traces} provides full recipe, chronic-health,
and FPS traces.

\noindent \textbf{Boundary leakage through semantic residue.}
In PersonaMem user \texttt{139}, the benchmark marks ``Reads African literature classics'' as do-not-remember. The tracer correctly learns the explicit boundary: the assistant should not assume familiarity with African classics or restrict recommendations to African authors/settings. However, the same turn also produces nearby positive hypotheses about discussion-oriented postcolonial literature, which are not marked as negative-only and survive consolidation into the final Books/Literature profile. The error is thus not stale memory, but same-turn semantic residue around a forbidden attribute. This motivates boundary-aware consolidation, where do-not-remember signals also suppress semantically adjacent positive memories.

\section{Conclusion}

We study online personalization for language models, where user preferences are latent and gradually revealed through interaction. Existing training-free methods often fail to model uncertainty over user intent or reconcile long-term preferences with short-term topic-specific needs. To address this, we propose \method, a training-free framework that formulates personalization as latent preference tracing. Experiments on PRISM and PersonaMem-v2 show stronger and more robust response alignment and preference prediction than the evaluated online baselines, together with competitive long-term profile alignment. This further suggests a broader principle for personalization: combining fine-grained short-term belief updates with long-term memory consolidation can support adaptation that is robust, generalizable, and interpretable.

\section*{Limitations}

\methodname assumes explicit comparative choices; it does not directly infer
preferences from verbal critiques or unobserved implicit behavior, although its
state remains useful when choice updates are intermittent. Extending the
observation model without losing inspectability is an important direction. Its
natural-language hypotheses also inherit the underlying LLM's limits in
granularity and reliability, motivating structured representations,
user-editable memory, and safety-aware verification.

\section*{Ethical Considerations}

\subsection*{Artifacts Usage}

We use public datasets \cite{kirk2024prism, personamemv2}, models
\cite{qwen3.5, kimiteam2026kimik2openagentic, glm5team2026glm5vibecodingagentic,
deepseekai2026deepseekv4}, and APIs \cite{singh2026openaigpt5card} under their
released terms. We do not redistribute source data or model weights, identify
users, or add personally identifying information; evaluation is aggregate and
research-only.

\subsection*{AI Usage}

We used AI assistants for writing, editing, and code debugging. The authors made
and verified the research decisions, analyses, and claims, and reviewed all
assisted text or code. AI was not used to generate experimental results.

\subsection*{Potential Risks}

Personalization can leak sensitive information, misprofile users, amplify bias,
or become overly persuasive. We limit exposure through controlled offline and
aggregate evaluation on public research data. Explicit external traces also
provide an interface for future inspection, editing, deletion, and retention
controls, although they do not by themselves remove these risks.

\bibliography{custom}

\appendix
\label{sec:appendix}

\section{Implementation Details}
\label{app:implementation-details}

All experiments use the tracing procedure in \S\ref{sec:intra_session_updating} and
the prompt family in Appendix~\ref{app:prompt-family}, with dataset-specific prompt adapters. The
tracer maintains $K=5$ active natural-language hypotheses and includes at most
three recent turns in LLM prompts. Skip gating is enabled by default: for each
turn, the gate is queried five times, and the turn is skipped only when the
majority vote indicates insufficient preference evidence. Before hypothesis
updates, candidate responses are compressed into indexed summaries while
preserving the original candidate order and chosen-vs.-rejected labels. The
Bradley--Terry temperature is set to 1.0.

Initialization produces exactly five topic-labelled hypotheses. On later usable
turns, each slot is independently marked for revision or replacement; if a
majority of slots are marked for replacement, the working belief is
reinitialized, otherwise each replacement remains in its original slot. After
weighting and resampling, exact duplicates are grouped directly and other
near-duplicates are grouped using embedding similarity threshold $\tau=0.8$.
Each non-singleton group $G$ is merged into one hypothesis and expanded with
$|G|-1$ proposals on axes not represented by the remaining particles, restoring
the five-slot belief.

For cross-session memory, long-term hypotheses are stored in a FAISS
inner-product vector index. We use \texttt{openai/text-embedding-3-small}
embeddings with 1536 dimensions. Topic metadata is embedded with the hypothesis
content by default and removed only in the no-topic ablation. At a session
boundary, the initial user message retrieves the five nearest items for the
initializer's reuse-or-create decision. At response time, the system retrieves
up to 30 long-term items, keeps at most 5 items above the stored-weight threshold of 0.1,
and combines them with the current working belief to synthesize the
response-time profile.

All structured LLM outputs are constrained by typed JSON schemas, and malformed
outputs are rejected. The main run uses \texttt{openai/gpt-5} through OpenRouter
for tracing, response generation, and prediction. Evaluation uses
\texttt{google/gemini-3-flash-preview}. The hybrid setting keeps the same
algorithm and prompt family, but routes selected substeps to cheaper models:
preprocessing, branching, merging, and summary generation use
\texttt{openai/gpt-5-mini}, while initialization, surrogate choice scoring,
perturbation, response generation, profile synthesis, and prediction use
\texttt{openai/gpt-5}.

\section{Prompt Family and Usage}
\label{app:prompt-family}

Our method uses a modular prompt family rather than a single monolithic prompt.
Each experiment loads the same set of algorithmic prompt slots through a
dataset-specific adapter. Across model variants, the prompt semantics are kept
fixed; variants differ only in model routing or system configuration.

For prompts whose outputs are consumed by the tracing algorithm, we enforce
typed JSON schemas and reject malformed outputs. At each online turn, the system
builds or retrieves a response-time profile, generates an adapted response, and
then updates its belief state from the observed chosen-vs.-rejected comparison.
Low-signal turns can be filtered by a skip gate. Usable turns are summarized
into candidate contrasts, used to initialize or revise preference hypotheses,
reweighted by surrogate choice scoring, and periodically summarized, consolidated, or
perturbed to maintain a diverse hypothesis store. Table~\ref{tab:prompt-slots}
summarizes the prompt slots used in this process.

\begin{table*}[t]
\centering
\small
\setlength{\tabcolsep}{4pt}
\caption{Prompt slots used by the tracing framework.}
\begin{tabularx}{\textwidth}{@{}p{0.19\textwidth}p{0.25\textwidth}X@{}}
\toprule
\textbf{Prompt slot} & \textbf{When used} & \textbf{Purpose and output} \\
\midrule
\texttt{skip} & Before online update & Decides whether the turn contains usable preference evidence; returns a reason and Boolean decision. \\
\texttt{preprocessing} & After a turn passes the gate & Identifies preference-relevant candidate contrasts and returns indexed summaries. \\
\texttt{initialization} & When no working belief is active & Creates exactly $K$ topic-labelled hypotheses, reusing retrieved items when appropriate. \\
\texttt{branching} & Later usable turns & Revises or replaces each hypothesis independently according to the new evidence. \\
\texttt{likelihood} & After initialization or branching & Scores candidate alignment under one hypothesis; the scores induce the surrogate Bradley--Terry choice score. \\
\texttt{axis}, \texttt{merge}, \texttt{perturb} & Particle rejuvenation & Extracts preference axes, merges near-duplicates, and restores diversity after collapse. \\
\texttt{consolidate} & Session boundaries or reinitialization & Merges related long-term items and updates the global hypothesis store. \\
\texttt{summary}, \texttt{profile} & Response-time and final profiles & Compiles selected hypotheses into concise, actionable preference descriptions. \\
\texttt{response} & Before observing the gold choice & Produces an adapted response without exposing the personalization process. \\
\texttt{prediction} & Preference-prediction evaluation & Ranks candidates using the inferred profile and recent history. \\
\texttt{response\_evaluation} & Response-alignment evaluation & Scores the adapted response against each observed candidate on shared dimensions. \\
\texttt{profile\_evaluation} & Profile evaluation & Compares the inferred profile with survey or memory evidence under dataset-specific rubrics. \\
\bottomrule
\end{tabularx}
\label{tab:prompt-slots}
\end{table*}

\paragraph{Dataset adapters.}
Dataset adapters specialize the same prompt slots to different supervision
formats. For survey-based preference data, the adapter emphasizes stable
conversation-level preferences, such as communication style, structure,
factuality expectations, safety boundaries, and helpfulness criteria, while
avoiding overfitting to incidental topic facts. Survey fields are treated as
partial ground truth: demographic information is used only as a weak
compatibility signal and is not used to infer preferences.

For memory-based data, the adapter represents hypotheses as typed memory units
covering background facts, domain preferences, preference updates, constraint
boundaries, ownership boundaries, and adaptation rules. It separates
user-owned preferences from unsupported or other-person cues, treats privacy
and do-not-remember evidence as negative constraints, and applies only
message-relevant cues at response time. Its profile judge therefore focuses on
preference coverage, personalization utility, update and boundary handling, and
memory quality.

For the flat-slot ablation, only branching changes. The system keeps a fixed
number of active hypothesis slots and rewrites irrelevant slots in place;
hierarchical retrieval and consolidation cannot repair stale hypotheses.

\section{Additional Tracing Sensitivities}
\label{app:additional-tracing-analyses}

This section complements the main robustness analysis with controls on
initialization, surrogate scoring, propagation, and particle rejuvenation.

\subsection{Initialization and Scoring Consistency}

\begin{table}[t]
\centering
\footnotesize
\setlength{\tabcolsep}{3.2pt}
\renewcommand{\arraystretch}{1.05}
\caption{Consistency checks on 100 fixed-hypothesis PRISM events. Cold-start
matches use independent one-to-one matching with Gemini-3-Flash. JSD values use
three scoring repeats per model; lower is more consistent.}
\textbf{Cold-start hypotheses}\\[1mm]
\begin{tabular}{lc}
\toprule
\textbf{Comparison} & \textbf{Matched hypotheses} \\
\midrule
GPT-5 / repeat & 4.43/5 \\
GPT-5 / GPT-5-mini & 4.07/5 \\
GPT-5 / Qwen-3.5-9B & 3.80/5 \\
\bottomrule
\end{tabular}

\vspace{2mm}
\textbf{Surrogate utility estimates}\\[1mm]
\begin{tabular}{lcc}
\toprule
\textbf{Comparison} & \textbf{Candidate JSD} & \textbf{Particle JSD} \\
\midrule
GPT-5 / repeat & 0.010 & 0.008 \\
GPT-5 / GPT-5-mini & 0.043 & 0.022 \\
GPT-5 / Qwen-3.5-9B & 0.044 & 0.025 \\
\bottomrule
\end{tabular}
\label{tab:initialization_scorer_consistency}
\end{table}

\autoref{tab:initialization_scorer_consistency} places the two consistency
checks together because both concern LLM sensitivity at fixed input. A second
GPT-5 initialization recovers 4.43 of five hypotheses on average; GPT-5-mini
and Qwen-3.5-9B recover 4.07 and 3.80. Repeated GPT-5 utility estimates have low
JSD, while the smaller alternative scorers produce related, though less
similar, distributions. Across 300 GPT-5 scoring passes, 270 yield non-uniform
support over the five hypotheses, with mean normalized Gini $0.168$. These are
descriptive consistency checks rather than evidence of score calibration.

\subsection{Particle Propagation}

\begin{table}[t]
\centering
\footnotesize
\setlength{\tabcolsep}{3.2pt}
\renewcommand{\arraystretch}{1.05}
\caption{Particle-propagation sensitivity on 28 matched PRISM users.}
\begin{tabular}{lccc}
\toprule
\textbf{Setting} & \textbf{Acc$_{>20}$} & \textbf{GPT$_{>20}$} & \textbf{Prof.} \\
\midrule
Standard hybrid & 0.5878 & 0.1219 & 4.2500 \\
Heterogeneous & 0.5617 & 0.1547 & 4.0714 \\
\bottomrule
\end{tabular}
\label{tab:heterogeneous_propagation}
\end{table}

In the heterogeneous condition in \autoref{tab:heterogeneous_propagation},
GLM-5.2 initializes the five particles, and Kimi-K2.6, GPT-5-mini,
Qwen-3.5-9B, GLM-5.2, and DeepSeek-V4-Flash independently propagate fixed
slots. Metrics include turns with at least 10 active users. GPT-5 still
performs utility scoring, and the reported online metrics are smoothed
post-turn-20 averages. This control therefore tests particle
generation rather than isolating the scorer. Preference prediction and profile
alignment decline, while response alignment increases.

\subsection{Axis-Based Rejuvenation}

\begin{table}[t]
\centering
\footnotesize
\setlength{\tabcolsep}{3.2pt}
\renewcommand{\arraystretch}{1.05}
\caption{Perturbation-rule sensitivity on the matched 28-user PRISM cohort,
using smoothed post-turn-20 averages with at least 10 active users.}
\begin{tabular}{lccc}
\toprule
\textbf{Rule} & \textbf{Acc$_{>20}$} & \textbf{GPT$_{>20}$} & \textbf{Prof.} \\
\midrule
Axis-based & 0.5878 & 0.1219 & 4.2500 \\
Rephrase-only & 0.5596 & -0.0075 & 4.1000 \\
\bottomrule
\end{tabular}
\label{tab:perturbation_sensitivity}
\end{table}

After grouping near-duplicates at similarity threshold $\tau=0.8$, the
standard procedure merges each group $G$ and requests $|G|-1$ proposals on
preference axes not already represented. Replacing this step with paraphrases
of the collapsed hypothesis reduces all three metrics in
\autoref{tab:perturbation_sensitivity}, favoring axis-level diversification
over surface variation.

\section{More Qualitative Examples}
\label{app:illustrative-traces}

Drawn from logged preprocessing and tracing outputs, the cases illustrate
five-slot candidate contrasts rather than provide an annotated evaluation.

\paragraph{Recipe request.}
The user asks for a recipe using tortillas, corn, canned salmon, beans, red
enchilada sauce, and green chiles. The selected candidate is a concise baked
recipe using the requested ingredients; alternatives are verbose, taco-style,
or unrelated. The trace separates five cues: (1) concise, direct instructions;
(2) baked, casserole-style preparation rather than stovetop tacos; (3) explicit
use of all listed pantry items; (4) brief serving suggestions alongside the
core recipe; and (5) rejection of off-topic or meandering content. Each slot
reflects visible candidate differences without unsupported assumptions about
the user.

\paragraph{Chronic-health support.}
The user describes fluctuating chronic illness and limited daily capacity. The
selected candidate is empathetic, personalized, and practical, whereas the
alternatives are more generic or formal. The five hypotheses capture a
validating tone, actionable follow-up, recognition of proactive health
management and a support network without patronizing language, collaborative
framing that lets the user define priorities, and sensitivity to good and bad
days. The trace remains at the level of response preferences and does not infer
new medical facts.

\paragraph{FPS refinement.}
The user likes Counter-Strike gunplay, dislikes team play, and prefers
deathmatch. From a selected recommendation with a concrete
individual-performance mode, the trace refines an earlier broad competitive-FPS
hypothesis into preferences for (1) realistic, skill-focused deathmatch or
arena modes, (2) low time-to-kill, (3) precise hitscan-centric gun models,
(4) minimal power-granting progression, and (5) strong anti-cheat and
competitive integrity. Here, later explicit rejections narrow a broad genre
preference into factors that can guide subsequent recommendations.

\section{Detailed Turn-Level Results}
\label{app:detailed-turn-level-results}

\begin{figure*}[t]
    \centering
    \begin{subfigure}[t]{0.485\textwidth}
        \centering
        \includegraphics[width=\linewidth]{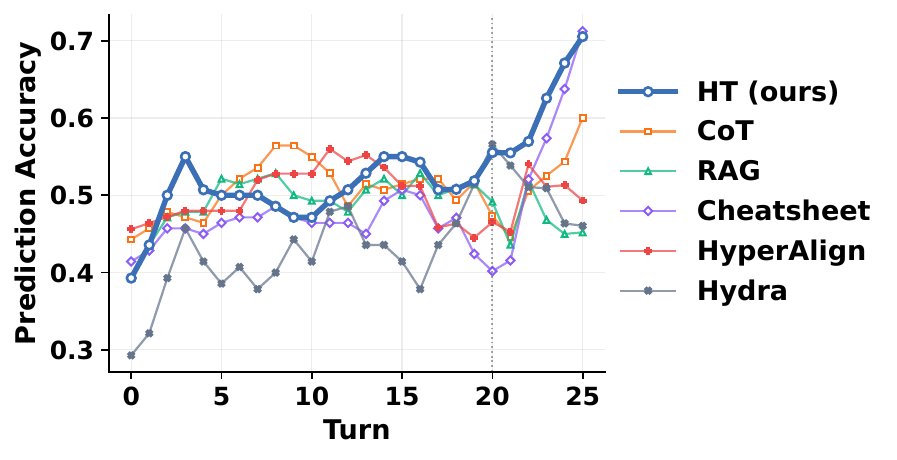}
        \caption{PRISM: Accuracy.}
    \end{subfigure}
    \hfill
    \begin{subfigure}[t]{0.485\textwidth}
        \centering
        \includegraphics[width=\linewidth]{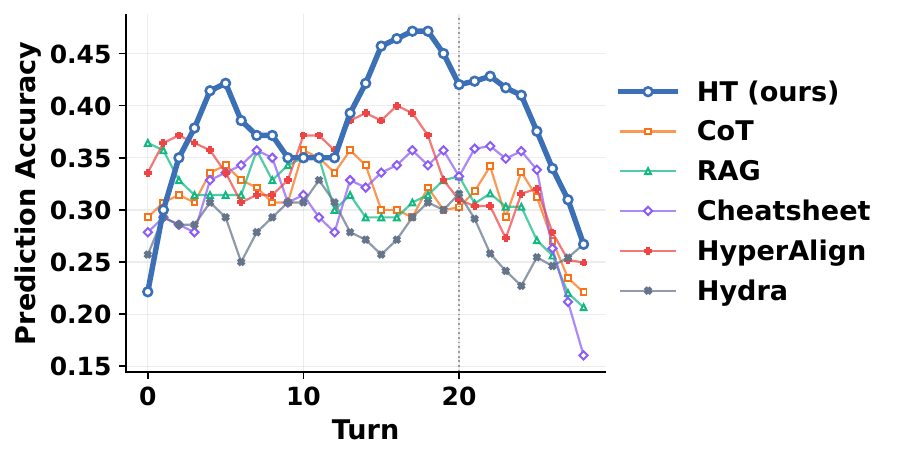}
        \caption{PersonaMem-v2: Accuracy.}
    \end{subfigure}

    \vspace{1.2mm}

    \begin{subfigure}[t]{0.485\textwidth}
        \centering
        \includegraphics[width=\linewidth]{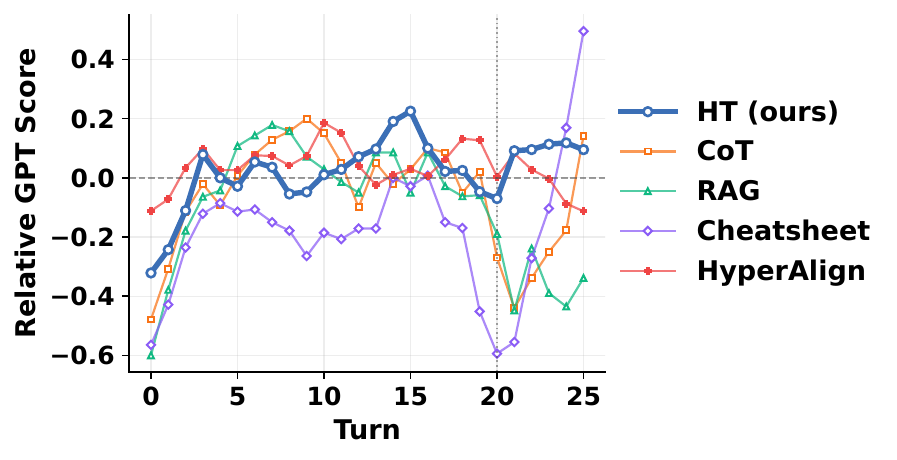}
        \caption{PRISM: Relative GPT score.}
    \end{subfigure}
    \hfill
    \begin{subfigure}[t]{0.485\textwidth}
        \centering
        \includegraphics[width=\linewidth]{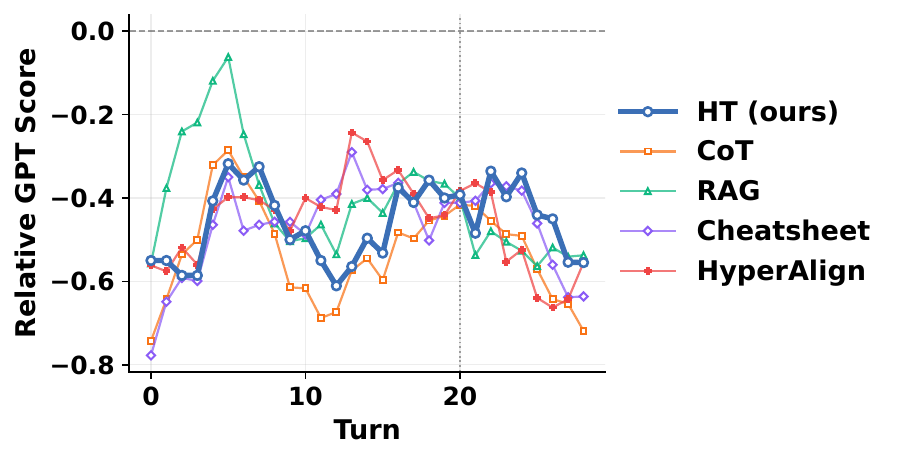}
        \caption{PersonaMem-v2: Relative GPT score.}
    \end{subfigure}

    \vspace{1.2mm}

    \begin{subfigure}[t]{0.485\textwidth}
        \centering
        \includegraphics[width=\linewidth]{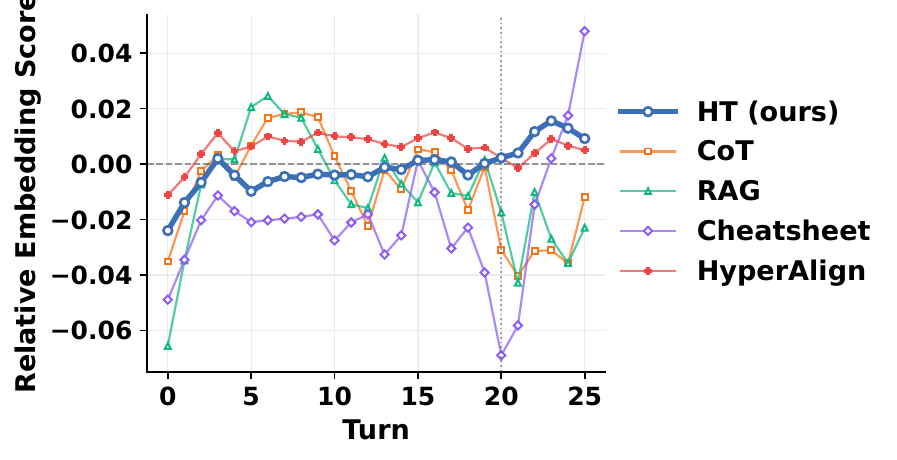}
        \caption{PRISM: Relative embedding score.}
    \end{subfigure}
    \hfill
    \begin{subfigure}[t]{0.485\textwidth}
        \centering
        \includegraphics[width=\linewidth]{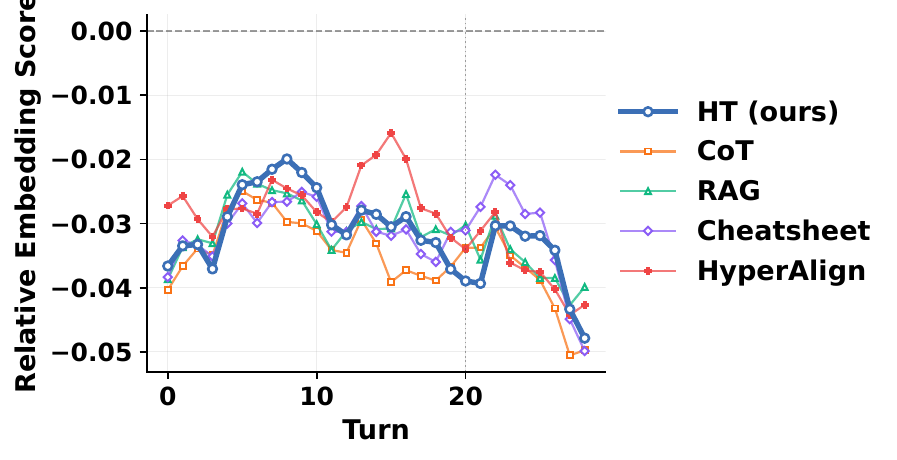}
        \caption{PersonaMem-v2: Relative embedding score.}
    \end{subfigure}

    \vspace{1.2mm}

    \begin{subfigure}[t]{0.485\textwidth}
        \centering
        \includegraphics[width=\linewidth]{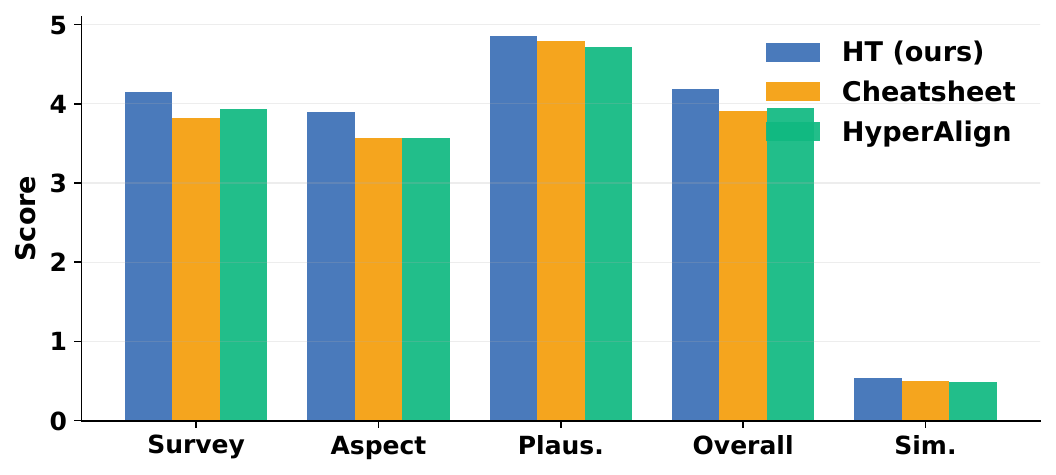}
        \caption{PRISM: Profile score.}
    \end{subfigure}
    \hfill
    \begin{subfigure}[t]{0.485\textwidth}
        \centering
        \includegraphics[width=\linewidth]{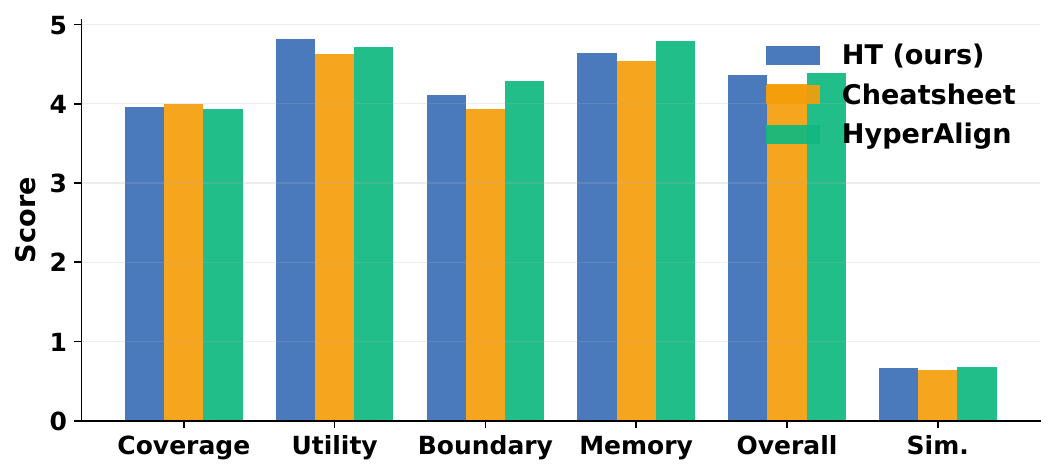}
        \caption{PersonaMem-v2: Profile score.}
    \end{subfigure}

    \caption{Detailed turn-level online evaluation results on PRISM and
    PersonaMem-v2. Curves are plotted at each interaction turn.}
    \label{fig:detailed_turn_level_results}
    \vspace{-2mm}
\end{figure*}

The main text reports the PRISM adaptation curves with SEM and summarizes both
datasets through aggregate statistics. Here, we provide the complete turn-level
curves for PRISM and PersonaMem-v2 in
\autoref{fig:detailed_turn_level_results}.

\end{document}